\pdfoutput=1

\documentclass[11pt]{article}

\usepackage[review]{acl}

\usepackage{times}
\usepackage{latexsym}
\usepackage[review]{acl}
\usepackage{times}
\usepackage{latexsym}
\usepackage{tipa}
\usepackage{amsmath}
\usepackage{amsfonts} 
\usepackage{booktabs}
\usepackage{multirow}
\usepackage{adjustbox}
\usepackage{longtable}
\usepackage{graphicx}
\usepackage{float} 
\usepackage[T1]{fontenc}

\usepackage[utf8]{inputenc}

\usepackage{microtype}

\usepackage{inconsolata}

\usepackage{adjustbox}
\usepackage{afterpage}

\title{Embedded Translations for Low-resource Automated Glossing}

\author{
Changbing Yang, Garrett Nicolai, Miikka Silfverberg \\
University of British Columbia \\
{\tt cyang33@@mail.ubc.ca \quad miikka.silfverberg@ubc.ca}
}

\begin{document}
\nolinenumbers
\maketitle
\begin{abstract}

We investigate automatic interlinear glossing in low-resource settings. We augment a hard-attentional neural model with embedded translation information extracted from interlinear glossed text. After encoding these translations using large language models, specifically BERT and T5, we introduce a character-level decoder for generating glossed output. Aided by these enhancements, our model demonstrates an average improvement of 3.97\%-points over the previous state of the art on datasets from the SIGMORPHON 2023 Shared Task on Interlinear Glossing. In a simulated ultra low-resource setting, trained on as few as 100 sentences, our system achieves an average 9.78\%-point improvement over the plain hard-attentional baseline. These results highlight the critical role of translation information in boosting the system's performance, especially in processing and interpreting modest data sources. Our findings suggest a promising avenue for the documentation and preservation of languages, with our experiments on shared task datasets indicating significant advancements over the existing state of the art.    
\end{abstract}

\section{Introduction}

The extinction rate of languages is alarmingly high, with an estimated 90\% of the world's languages at risk of disappearing within the next century~\cite{krauss1992world}. As speaker populations decline, linguists are urgently prioritizing the documentation of these languages. This documentation process is multi-faceted, involving phonetic and orthographic transcription, translation, morpheme segmentation, and linguistic annotation~\cite{crowley2007field}. This information is traditionally represented as Interlinear Glossed Texts (IGT). An example for the Tsimshianic language Gitksan is given below (see Appendix \ref{sec:appendix} for additional details):
\begin{adjustbox}{width=\columnwidth}
    \begin{tabular}{ll}
    & \\
        \textbf{Orthography:} & Ii hahla'lsdi'y goohl IBM \\
        \textbf{Segmentation:} & ii hahla'lst-'y goo-hl IBM \\
        \textbf{Gloss:} & CCNJ work-1SG.II LOC-CN IBM \\
        \textbf{\color{red} Translation:} & \color{red} And I worked for IBM.\\
        &  \\
    \end{tabular}
\end{adjustbox}


The traditional manual approach to language documentation, while thorough, is notably labor-intensive. This has spurred the development of automated tools leveraging machine learning for tasks such as word segmentation and glossing. 


\begin{figure}[H]
  \includegraphics[width=0.5\textwidth]{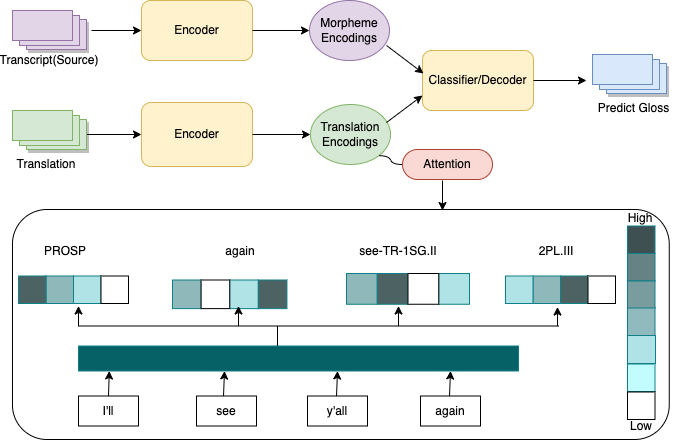}
  \caption{Pipeline of the proposed work. The lower portion of the diagram demonstrates how attention weights inform the model when predicting the glossing targets.}
  \label{fig:pipeline2}
\end{figure}

Our glossing system\footnote{Our code is publicly available: \url{https://github.com/changbingY/Auto_glossing_stem_translation}} is based on the winning submission \cite{girrbach2023tu} for the SIGMORPHON 2023 shared task on interlinear glossing \cite{ginn2023findings}.
To this model, we incorporate additional supervision in the form of translations in a high-resourced language like English or Spanish, (see the {\bf\color{red} Translation} line in the IGT example above). Utterances are typically manually translated during the language documentation process. Compared to manual glossing, translation is a fast operation, which makes it well-suited for use as an additional cheap source of supervision. 

We experiment with incorporating utterance translations (already present in SIGMORPHON shared task data) using different encoders: a vanilla LSTM \cite{hochreiter1997long}, BERT \cite{kenton2019bert} and T5 \cite{raffel2020exploring}. The latter two are pre-trained for the translation target language. We investigate different ways to incorporate encoded translation information into the glossing model and show that soft attention \cite{bahdanau2014neural} over translation token representations delivers the best performance. In addition to the  translation encoder, we add a character-based decoder to the model, which is particularly helpful in low-resource settings. Together, these enhancements lead to a substantial 3.97\%-point improvement in glossing accuracy over the strong \newcite{girrbach2023tu} model on SIGMORPHON shared task datasets and a 9.78\%-point improvement  improvement in an ultra low-resource setting. 



 
We are not unique in incorporating translation information into a glossing system. The system presented by \newcite{okabe-yvon-2023-towards} is based on CRFs \cite{sutton2012introduction}, and also employs 
translations. However, in contrast to our approach, they heavily rely on source and target word alignments derived from an unsupervised alignment system \cite{jalili-sabet-etal-2020-simalign}. In low-resource settings, it is hard to learn an accurate alignment model.
\footnote{Moreover, \newcite{okabe-yvon-2023-towards} assume morphologically-segmented input, which considerably simplifies the glossing task. We instead address the much harder task of predicting glosses without segmentation information.} 

Pioneering studies by \newcite{zoph2016multi}, \newcite{anastasopoulos2018leveraging} and \newcite{zhao-etal-2020-automatic}, show that leveraging translations can enhance the performance of a neural glossing system. 
A notable limitation in all of these approaches is the scarcity of available English translations for training models. Therefore, only modest improvements in glossing accuracy are observed. Our work, in contrast, incorporates translation information through large pre-trained language models, which leads to greater improvements in glossing performance. 
This strategy has lately become increasingly popular in low-resource NLP and 
shows promise across various language processing tasks \cite{ogueji2021small, hangya2022improving}.

Similarly to our approach, \newcite{okabe-yvon-2023-towards} also take advantage of the BERT model in their study, but only utilize BERT representations for translation alignment. In contrast, we directly incorporate encoded translations into our glossing model. 
\newcite{he2023sigmorefun} also use pre-trained language models, namely, XLM-Roberta \cite{conneau2020unsupervised}, mT5 \cite{xue2021mt5} and ByT5 \cite{xue2022byt5}, as part of their glossing model. However, they do not incorporate IGT translation information.\footnote{Though \newcite{he2023sigmorefun} do use external dictionary information for post-correction of glosses.} Instead, they directly fine-tune the pre-trained models for glossing. 

Our key-contributions are: 
{\bf 1.} We apply large pre-trained language models to incorporate translation data into the automatic glossing process. 
{\bf 2.} We analyze attention distributions over encoded translation tokens, showing that our models derive useful knowledge from translations.
{\bf 3.} We substantially aid peformance in low-resource settings by introducing a character-based decoder.


\section{Experiments}
\label{experiment}
We present experiments on interlinear glossing. As input, our model uses a sentence like {\it Le chien aboie} (Fr.) and its translation `The dog barks'. It then produces a sequence of glossed tokens as output:  {\bf ART dog bark-IND.PRES.3SG} (one space-delimited token-gloss per input token). We conduct experiments on data from the 2023 SIGMORPHON shared task on interlinear glossing \cite{ginn2023findings} which encompasses six languages: Arapaho, Gitksan, Lezgi, Nat\"ugu, Tsez and Uspanteko.\footnote{The datasets are further described in Appendix \ref{appendix-dataset}.} 
In order to investigate the performance of our model in ultra low-resource settings, we additionally form smaller training sets by sampling 100 sentences from the original shared task training data.\footnote{We do this for all shared task languages apart from Gitksan, which only has 30 training sentences.} In both settings, we use the original shared task development and test sets for validation and testing, respectively. 
In all experiments, we evaluate based on
token-level glossing accuracy and edit distance.  

\textbf{Baseline Model} Our research builds upon the hard-attentional glossing model developed by \newcite{girrbach2023tu} which won the 2023 SIGMORPHON shared task. 
Glossing is performed through three stages: \textbf{1. Orthographic Input Encoding}: The model first encodes the input utilizing a character-based BiLSTM encoder. \textbf{2. Morpheme Segmentation}: The second phase involves an unsupervised segmentation process, which relies on the encoded character-level embeddings to discern individual morphemes. \textbf{3. Morpheme Classification}: The model employs a linear classifier to predict glosses for the previously segmented morphemes. 
It is important to note that, due to employing a simple classifier, the model is constrained to generating glosses observed in the training data; it can, for example, never generate an unseen stem gloss. 
We address this limitation by integrating a character-based decoder as discussed below.

\textbf{Translation Information} We extend the model of \newcite{girrbach2023tu} by incorporating translations. We encode the English or Spanish (in the case of Uspanteko) translations in the shared task datasets using a deep encoder: a randomly-initialized character-based BiLSTM or pre-trained BERT-base/T5-large.\footnote{See Appendix \ref{model_setting} for details concerning the encoders.} To represent translations, we then either use the final hidden state from the translation encoder, or attend over the translation hidden states. In both cases, translation information is concatenated with the encoder hidden state and fed to the decoder.
To compute the attention weights for decoder state $d_i$, we extract encoder states $(e_j)_{j=1}^{J}$, where $J$ is the number of morphemes in the input\footnote{Morphemes are discovered by the \newcite{girrbach2023tu} model in an unsupervised manner during training.}, and translation encoding vectors $(t)_{k=1}^{K}$, where $K$ is the number of subwords (or characters when using the BiLSTM encoder) in the translation. We then apply Bahdanau attention \cite{bahdanau2014neural} to $d_i$ and concatenated vectors $[e_j; t_k]$.\footnote{There are $J \times K$ of these vectors, in total.}

\textbf{Character-Based Decoder} Our second addition to the \newcite{girrbach2023tu} model is a character-based decoder. As noted above, the baseline model is unable to predict glosses which were not present in the training data. This deficiency is particularly harmful when predicting glosses for lexical morphemes (i.e. word stems) which represent a much larger inventory than grammatical morphemes (i.e. inflectional and derivational affixes). A character-based decoder can enhance the model's capability to generate glosses, for example, by learning to copy lexical morphemes from the translation to the gloss.
Following \newcite{kann2016med}, the decoder is implemented as an LSTM.

\textbf{Details} We train ten parallel models and use majority voting to predict the gloss for each token. 




\section{Results}
\begin{table*}[]
\begin{adjustbox}{width=\textwidth,center}
\begin{tabular}{lllllll|lllllll}
\toprule
Model setting        & arp  & lez  & ntu  & ddo & usp & ave&  arp-low  & git-low & lez-low  & ntu-low   & ddo-low  & usp-low &ave      \\\midrule
\newcite{girrbach2023tu} & 78.79& 78.78  & 81.04  & 80.96 & 73.39 & 78.59 &19.12 & 21.09 & 48.84  & 51.08 & 36.12 & 17.32 & 32.26 \\\bottomrule
LSTM & 77.04& 81.42 & 83.55 & 84.99 & 73.01 & 80.00&18.67 & 20.71 &54.29  & 59.56 & \textbf{44.5}  & 32.92 & 38.44\\
LSTM+attn & 79.31 & 76.19 & 83.01 & 85.12 & 76.24 &79.97 &24.38 &  18.49 &  55.75 & 58.48 & 42.37 & 29.52 & 38.17\\
BERT+attn & 78.98 & 81.87 & 84.57 & 85.84 & 77.63 &81.78 & 27.33 &  20.31  & 55.86  & 60.13  & 41.85 & 33.04 & 39.75\\
BERT+attn+chr & 80.79 & 82.19 & \textbf{85.41} & 84.13 & \textbf{79.34}   & 82.37&\textbf{28.82} & \textbf{28.11} & 56.99 & 62.73  & 39.72 & \textbf{35.84} & \textbf{42.04}   \\
T5+attn+chr & \textbf{81.11}  & \textbf{82.37} & 84.68 & \textbf{85.91} & 78.72 & \textbf{82.56} & 27.31  & 24.23 & \textbf{57.33} & \textbf{62.82} & 39.97 & 33.59& 40.88 \\\bottomrule
\end{tabular}
\end{adjustbox}
\caption{Word-level accuracy of languages in the 2023 Sigmorphon Shared Task \cite{ginn2023findings} (left) and low-resource settings (right), with `arp' representing Arapaho, `git' for Gitksan, 
 `lez' for Lezgi, `ntu' for Nat\"ugu, `ddo' for Tsez, and `usp' for Uspanteko. Model specifics are elaborated in Section \ref{experiment}. }
\label{app:gloss_stats1}
\end{table*}

Table \ref{app:gloss_stats1} shows the glossing accuracy across different model settings and languages.\footnote{We additionally present edit distance  in Appendix \ref{app:edit-distance}.} We report performance separately for original shared task datasets and our simulated ultra low-resource datasets spanning 100 training sentences. We group the Gitksan shared task dataset in the low-resource category because it only has 30 training examples.\footnote{Apart from the baseline, all models apply majority voting. Its impact is discussed in  Appendix \ref{majorityvoting}.}
We use 
\newcite{girrbach2023tu} as our baseline system. 

\textbf{Shared Task Data} When integrating translations through the final state of a randomly-initialized bidirectional LSTM, we observe an improvement in average glossing accuracy, but performance is reduced for two languages (Arapaho and Uspanteko). 
Augmenting translations via an attentional mechanism (LSTM+attn) still does not confer consistent improvements. 
In contrast, translation information incorporated via a pre-trained model (BERT+attn) renders consistent improvements in glossing accuracy across all languages and we see notable gains in average glossing accuracy over the baseline.  
Incorporating a character-based decoder leads to further improvements in average glossing accuracy and for all individual languages. The T5 model (T5+attn+chr) attains the highest average performance: 82.56\%, which represents a 3.97\%-point improvement over the baseline. It also  delivers the highest performance for three out of our five test languages (Arapaho, Lezgi and Tsez), while the BERT-based model and attention (BERT+attn+chr) delivers the best performance for the remaining two (Nat\"ugu and Uspanteko). 
Among all languages, we see improvements over \newcite{girrbach2023tu} ranging from 2.32\%-points to 5.95\%-points. 

\textbf{Ultra Low-Resource Data} 
Translations integrated through the final state of a randomly initialized bidirectional LSTM  (LSTM and LSTM+attn),
lead to an average 6\%-point improvement in accuracy over the baseline. We achieve particularly impressive gains for
Uspanteko, surpassing the baseline accuracy by over 15\%-points. 
Incorporating pre-trained models (BERT+attn) 
exhibits a slight increase in accuracy for certain languages. 
However, when we incorporate both pre-trained models and the character-based decoder (BERT+attn+chr and T5+attn+chr), we see larger gains in accuracy
across the board. Here, BERT achieves the highest average accuracy of 42.02\%, which represents a 9.78\%-point improvement over the baseline. It achieves the highest performance for three languages (Arapaho, Gitksan and Uspanteko), while T5 delivers the best performance for two of the languages (Lezgi and Nat\"ugu). The plain LSTM model attains the best performance for Tsez.


\section{Findings and Analysis}
\textbf{Translation Information and the Character-Level Decoder} Across both standard and low-resource settings, translations and the character-level decoder confer consistent improvements in glossing accuracy. Improvements are particularly notable in the ultra low-resource setting, where we achieve a 9.78\%-point improvement over the baseline.

\textbf{Majority Voting} 
Results presented in Appendix \ref{majorityvoting} demonstrate that, while majority voting improves accuracy, we achieve improvements over the baseline even without it.

\begin{figure}[h]
  \begin{minipage}{0.5\textwidth}
    \centering
    \includegraphics[width=\textwidth]{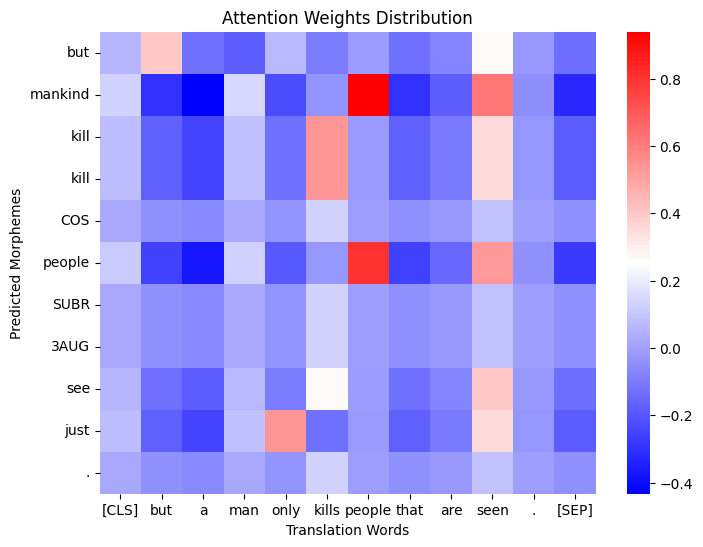}
  
  \end{minipage}
  \caption{Difference from mean attention weights of glossed output tokens (y-axis) with respect to encoded translation tokens (x-axis) for a Nat\"ugu example (attention weights are derived from the model BERT+attn+chr).}

  \label{fig:attdis-natugu-bert1}
\end{figure}

\textbf{Attention Distribution Visualization} To assess whether our model is able to successfully incorporate translation information, 
we visualize the attention patterns (from the BERT+attn+chr model) over the English translation representations. Figure \ref{fig:attdis-natugu-bert1} presents an example for Nat\"ugu. Attention weights are displayed in a heat map, where each cell indicates  difference from mean attention: $a - 1/(n+2)$. Here $n$ is the length of the translation in tokens ($+2$ here because of the start-of-sequence and end-of-sequence tokens \textsc{[cls]} and \textsc{[sep]} which 
are concatenated to the translation). 
 Positive red cells inidicate high attention and negative blue cells low attention. The visualization clearly indicates that the model attends to the relevant tokens in the translation when predicting the stems {\it people}, {\it mankind} and {\it kill}. Appendix \ref{attn-distribution-graphs-appendix} shows randomly picked heat maps for the rest of the languages. We can see that attention weights for the larger shared task datasets tend to express relevant associations, while attention weights for the ultra low-resource training sets largely represent noise. Appendix \ref{attn-distribution-graphs-appendix} also displays attention distributions when translations are incorporated using a randomly initialized LSTM instead of a pre-trained language model. These distributions also largely represent noise indicating that pre-trained models confer an advantage.


\section{Conclusions}
The study demonstrates the effectiveness of incorporating translation information and large pre-trained language models in automatic glossing for low-resource languages. The proposed system, based on a modified version of \newcite{girrbach2023tu}'s model, shows performance enhancements, particularly in low-resource settings. This research offers a potential efficient solution for aiding in the preservation of endangered languages. To sum up, this research contributes to the field of NLP by demonstrating the potential for achieving higher accuracy in language processing tasks, especially in linguistically-diverse and data-sparse environments.

\section{Limitations}
The limitations of our study primarily pertain to the extent of our experimentation and the models we have chosen. Firstly, our investigation relies solely on an LSTM decoder. This decision was influenced by time constraints, which limited our ability to explore more complex decoders. Additionally, our experimentation is confined to the T5-large model. While this model has shown promising results in our study, we acknowledge the existence of other large language models in the field of natural language processing. Although we did explore other large language models such as Llama2 \cite{touvron2023llama}, our preliminary experiments yielded unsatisfactory results compared to T5. Consequently, we made the decision not to include Llama2 in our paper due to its inferior performance. These limitations underscore the need for future research to explore a wider range of decoding architectures and incorporate various large language models to enhance our understanding of the subject matter. However, training large language models requires significant computational resources, which can have an environmental impact due to increased energy consumption. 

\bibliography{custom,anthology}


\appendix
\section{Appendix}
\subsection{IGT Information}
\label{sec:appendix}
In the IGT data, the second line includes segmentations with morphemes normalized to a canonical orthographic form. The third line has an abbreviated gloss for each segmented morpheme. Lexical morphemes typically correspond to the stems of words. The morpheme glosses usually have two categories: Lexical and Grammatical morphemes. For example, in glossing labels such as work-1SG.II, ``work" would be considered a Lexical morpheme, representing the core semantic unit. On the other hand, Grammatical morphemes like `1SG.II" are often denoted by uppercase glosses and generally signify grammatical functions, such as tense, aspect, or case, rather than specific lexical content.

\subsection{Shared Task Dataset}
\label{appendix-dataset}
This dataset is enriched with additional linguistic attributes such as morphological segmentations, varying according to the track—closed or open—thus offering a robust foundation for both training and evaluating linguistic models. Our analysis primarily focuses on data from Track 1, which emphasizes the use of transcription and translation information. The languages in this track include Arapaho, Gitksan, Lezgi, Nat\"ugu, Tsez, and Uspanteko. A notable aspect of this dataset is its resource constraints, with most languages, except Arapaho, comprising fewer than 10,000 sentences. This characteristic marks them as low-resource languages. In the case of Nyangbo, the dataset lacks translation information, leading us to exclude this language from our experiments. Additionally, it's worth mentioning that for Uspanteko, the provided translations are in Spanish, not English. Details of dataset information are shown as in Table \ref{tab:data-detail}.

\begin{table*}[h]
\centering
\begin{tabular}{|l|c|c|c|c|}
\hline
Language & Train sents & Dev sents & Test sents & Translations \\
\hline
Arapaho (arp) & 39,501 & 4,938 & 4,892 & (eng) \\
Gitksan (git) & 31 & 42 & 37 & (eng) \\
Lezgi (lez) & 701 & 88 & 87 & (eng) \\
Nat\"ugu (ntu) & 791 & 99 & 99 & (eng) \\
Tsez (ddo) & 3,558 & 445 & 445 & (eng) \\
Uspanteko (usp) & 9,774 & 232 & 633 & (spa) \\
\hline
\end{tabular}
\caption{2023 Sigmorphon Shared Task Dataset Information \cite{ginn2023findings}.}
\label{tab:data-detail}
\end{table*}

\subsection{Edit Distance}
\label{app:edit-distance}
Results are shown in Table \ref{app:gloss_stats_edit_distance}.
\begin{table*}[]
\begin{adjustbox}{width=\textwidth,center}
\begin{tabular}{lllllll|lllll}
\toprule
Model setting        & ara & git(-low)  & lez  & ntu  & ddo & usp &  ara-low   & lez-low  & ntu-low   & ddo-low  & usp-low       \\\midrule
\newcite{girrbach2023tu} & -&- &- &- &- &- &6.59   &3.64  & 4.78 & 4.92 & 3.79\\
LSTM & 1.52 & 5.65  &1.22  & \textbf{1.17 } & 0.72 & 0.88 &6.50& 3.28 & 4.12 & \textbf{3.93}  &  2.84 \\
LSTM+attn & \textbf{1.31}& 6.27  &  1.62&  1.34& 0.72 & 0.86 & 6.04   & 3.26 & 3.81 & 4.25 &3.21  \\
BERT+attn & 1.39 & 5.57  & 1.24 &  1.23&  0.69&  \textbf{0.70} & 5.97 &3.20  & 3.81  &  4.1& 2.88\\
BERT+attn+chr & 1.50& \textbf{5.30} & 1.20 & 1.25 & 0.53 & 0.81 &\textbf{5.54} & 3.04  & \textbf{3.55}   & 4.27 & 2.78  \\
T5+attn+chr &1.40 & 5.51 & \textbf{1.18} & 1.27 & \textbf{0.52}  & 0.78 &5.62 & \textbf{3.00}  & \textbf{3.55}   & 4.36  & \textbf{2.74}  \\\bottomrule
\end{tabular}
\end{adjustbox}
\caption{Word-level edit distance of languages in the 2023 Sigmorphon Shared Task \cite{ginn2023findings} (left) and low-resource settings (right), with `arp' representing Arapaho, `git' for Gitksan, `lez' for Lezgi, `ntu' for Nat\"ugu, `ddo' for Tsez, and `usp' for Uspanteko. Model specifics are elaborated in Section \ref{experiment}. }
\label{app:gloss_stats_edit_distance}
\end{table*}

\subsection{Model Settings}
\label{model_setting}
Our experimental framework and hyperparameters draw inspiration from Girrbach's methodology, with a focus on organizing and optimizing the technical setup. For model optimization, we employ the AdamW optimizer \cite{loshchilov2017decoupled}, excluding weight decay, and set the learning rate at 0.001. Except for this specific adjustment, we maintain PyTorch's default settings for all other parameters.

Our configuration is structured to allow a range of experiments, varying from 1 to 2 LSTM layers, with hidden sizes spanning from 64 to 512, and dropout rates fluctuating between 0.0 and 0.5. The scheduler $\gamma$ is adjusted within a range of 0.9 to 1.0, and batch sizes are diversified, ranging from 2 to 64. This versatile approach is designed to thoroughly evaluate the model's performance across a spectrum of hyperparameter configurations.

Departing from the original model which was trained for 25 epochs, our approach extends the training duration to 300 epochs when using large pretrained models. In cases where the BERT model is utilized, we sometime apply a 0.5 dropout rate during the BERT training phase. We exclusively employ the multilingual BERT model for Uspanteko, while we utilize the standard BERT model for all other languages. This comprehensive and meticulously organized setup is aimed at enhancing the effectiveness and efficiency of our model training process. 

To prevent coincidences, for each proposed model configuration, we train the model for 10 iterations, and the final prediction is determined through majority voting.

\subsection{Influence of Majority Voting}
\label{majorityvoting}
Average accuracy across 10 models and results utilized majority voting are shown in Table \ref{app:majority}. Improvements in performance can be achieved even without resorting to voting, particularly accentuated in ultra low-resource datasets as opposed to the Shared Task datasets.
\begin{table*}[]
\begin{adjustbox}{width=\textwidth,center}
\begin{tabular}{lllllll|lllllll}
\toprule
Model setting        & arp  & lez  & ntu  & ddo & usp & ave&  arp-low  & git-low & lez-low  & ntu-low   & ddo-low  & usp-low &ave      \\\midrule
\newcite{girrbach2023tu} & 78.79& 78.78  & 81.04  & 80.96 & 73.39 & 78.59 &19.12 & 21.09 & 48.84  & 51.08 & 36.12 & 17.32 & 32.26 \\\bottomrule

BERT/T5+attn+chr-average & 79.32 & 79.49 & 80.76 & 81.00 & 74.92  & 79.10 &25.43 & 23.95 & 54.28& 57.18 & 32.41 & 28.77 &  37.00 \\
BERT/T5+attn+chr-majority & \textbf{81.11}  & \textbf{82.37} & \textbf{85.41} & \textbf{85.91} & \textbf{79.34} & \textbf{82.83} &\textbf{28.82}  &  \textbf{28.11} & \textbf{57.33} & \textbf{62.82} & \textbf{39.97} &\textbf{35.84} &  \textbf{42.14} \\\bottomrule
\end{tabular}
\end{adjustbox}
\caption{Word-level accuracy of languages in the 2023 Sigmorphon Shared Task \cite{ginn2023findings} and low-resource settings. We compute the average across 10 models and also utilized majority voting accuracy results. Language abbreviations were used, with `arp' representing Arapaho, `git' for Gitksan, 
 `lez' for Lezgi, `ntu' for Nat\"ugu, `ddo' for Tsez, and `usp' for Uspanteko. Model specifics are elaborated in Section \ref{experiment}. }
\label{app:majority}
\end{table*}

\subsection{Attention Distribution}
\label{attn-distribution-graphs-appendix}

\begin{figure*}[htbp]
    \centering
    \includegraphics[width=\textwidth]{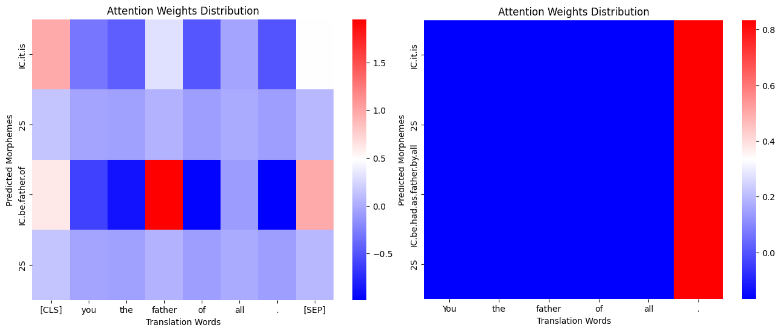}
    \caption{Difference from mean attention weights of glossed output tokens (y-axis) with respect to encoded translation tokens (x-axis) for an Arapaho example (attention weights are derived from the model BERT+attn+chr (left) and the model LSTM+attm (right)). The gold-standard glosses for this sentence: IC.it.is-2S IC.be.had.as.father.by.all-2S. }
    \label{fig:arpexample}
\end{figure*}

\begin{figure*}[htbp]
    \centering
    \includegraphics[width=\textwidth]{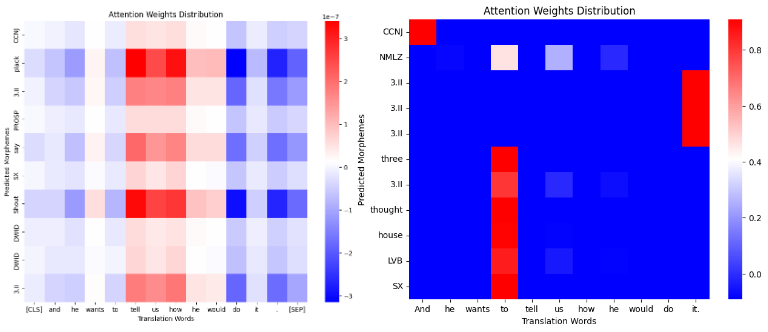}
    \caption{Difference from mean attention weights of glossed output tokens (y-axis) with respect to encoded translation tokens (x-axis) for a Gitksan example (attention weights are derived from the model BERT+attn+chr (left) and the model LSTM+attm (right)). The gold-standard glosses for this sentence: CCNJ want-3.II PROSP-3.I tell-T-3.II OBL-1PL.II MANR LVB-3.II.}
    \label{fig:gitexample}
\end{figure*}

\begin{figure*}[htbp]
    \centering
    \includegraphics[width=\textwidth]{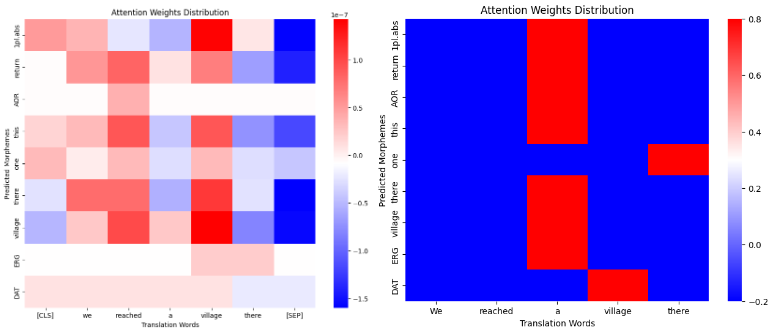}
    \caption{Difference from mean attention weights of glossed output tokens (y-axis) with respect to encoded translation tokens (x-axis) for a Lezgi example (attention weights are derived from the model BERT+attn+chr (left) and the model LSTM+attm (right)). The gold-standard glosses for this sentence: 1pl.abs return-AOR this one there village-ERG-DAT. }
    \label{fig:lezexample}
\end{figure*}

\begin{figure*}[htbp]
    \centering
    \includegraphics[width=\textwidth]{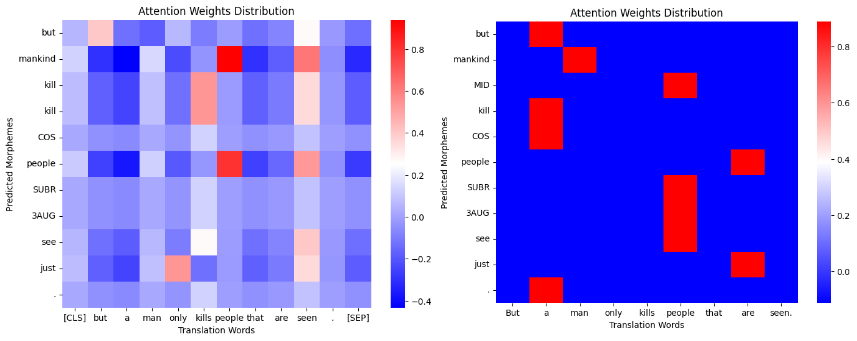}
    \caption{Difference from mean attention weights of glossed output tokens (y-axis) with respect to encoded translation tokens (x-axis) for a Nat\"ugu example (attention weights are derived from the model BERT+attn+chr (left) and the model LSTM+attm (right)). The gold-standard glosses for this sentence: but mankind MID-kill-COS-3MINIS people SUBR PAS-see-INTS-just.}
    \label{fig:ntuexample}
\end{figure*}

\begin{figure*}[htbp]
    \centering
    \includegraphics[width=\textwidth]{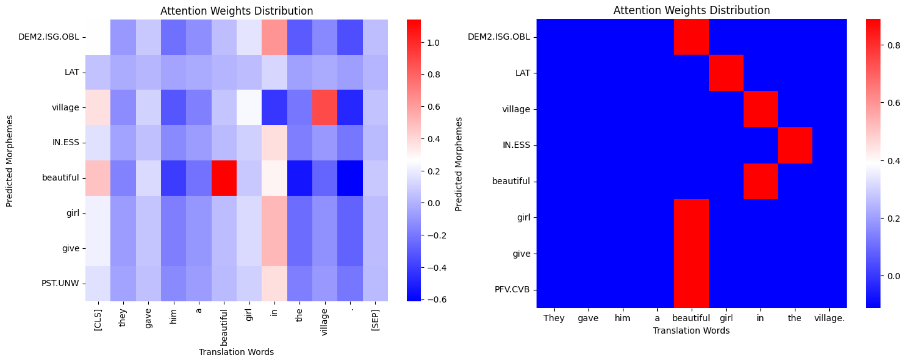}
    \caption{Difference from mean attention weights of glossed output tokens (y-axis) with respect to encoded translation tokens (x-axis) for a Tsez example (attention weights are derived from the model BERT+attn+chr (left) and the model LSTM+attm (right)). The gold-standard glosses for this sentence: DEM2.ISG.OBL-LAT village-IN.ESS beautiful girl give-PST.UNW}
    \label{fig:ddoexample}
\end{figure*}

\begin{figure*}[htbp]
    \centering
    \includegraphics[width=\textwidth]{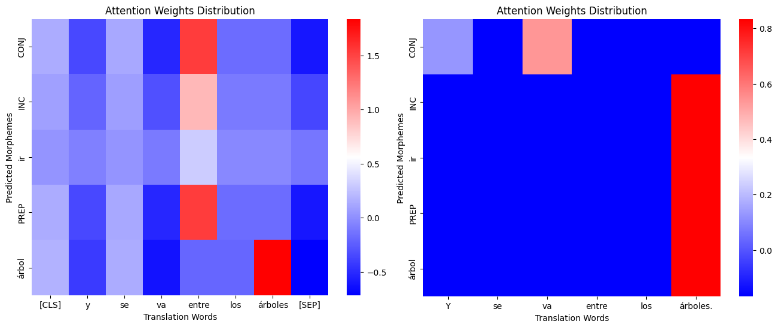}
    \caption{Difference from mean attention weights of glossed output tokens (y-axis) with respect to encoded translation tokens (x-axis) for a Uspanteko example (attention weights are derived from the model BERT+attn+chr (left) and the model LSTM+attm (right)). The gold-standard glosses for this sentence: CONJ INC-ir PREP árbol.}
    \label{fig:uspexample}
\end{figure*}

Attention distribution heat maps are shown in Figure \ref{fig:arpexample}-Figure \ref{fig:uspexample}.

\end{document}